\documentclass[sigplan, review=false, screen=true, balance=true]{acmart}

\renewcommand\footnotetextcopyrightpermission[1]{} 
\usepackage{booktabs} 
\usepackage{listings}
\usepackage{color}
\PassOptionsToPackage{hyphens}{url}\usepackage{hyperref}

\definecolor{dkgreen}{rgb}{0,0.6,0}
\definecolor{gray}{rgb}{0.5,0.5,0.5}
\definecolor{mauve}{rgb}{0.58,0,0.82}

\setcopyright{none}

\begin{document}

\title{A parallel Fortran framework for neural networks and deep learning}

\author{Milan Curcic}
\orcid{0000-0002-8822-7749}
\affiliation{%
  \institution{University of Miami}
  \streetaddress{4600 Rickenbacker Causeway}
  \city{Miami}
  \state{FL}
  \postcode{33149}
}
\email{mcurcic@miami.edu}

\begin{abstract}
This paper describes neural-fortran, a parallel Fortran framework for neural
networks and deep learning. It features a simple interface to construct
feed-forward neural networks of arbitrary structure and size, several activation
functions, and stochastic gradient descent as the default optimization algorithm.
neural-fortran also leverages the Fortran 2018 standard collective subroutines
to achieve data-based parallelism on shared- or distributed-memory machines.
First, I describe the implementation of neural networks with Fortran derived
types, whole-array arithmetic, and collective sum and broadcast operations
to achieve parallelism. Second, I demonstrate the use of neural-fortran in an
example of recognizing hand-written digits from images. Finally, I evaluate
the computational performance in both serial and parallel modes. Ease of use and
computational performance are similar to an existing popular machine learning
framework, making neural-fortran a viable candidate for further development and
use in production.
\end{abstract}


\maketitle

\section{Introduction}

Machine learning has seen tremendous increase in real-life applications
over the past two decades, with rapid development of algorithms, software,
and hardware to enable these applications. Some examples include image
recognition \citep{krizhevsky09}, natural language processing \citep{goldberg16},
and prediction of non-linear systems such as markets \citep{kimoto90} or
turbulent fluid flows \citep{kutz17}.
In contrast to classical computer programming where the input data is processed
with a predefined set of instructions to produce the output data, machine
learning aims to infer the instructions from sets of input and output data.
There are several widely used and battle-tested machine learning frameworks
available such as Tensorflow \citep{abadi16}, Keras \citep{chollet15},
PyTorch \citep{paszke17}, or Scikit-learn \citep{pendregosa11}.
Most high-level frameworks tend to be written in Python, with computational
kernels written in C++. With Fortran being a high-performance, compiled, and
natively parallel programming language, it is worth exploring whether Fortran
is suitable for machine learning applications.

Why consider Fortran as high-level language for machine learning applications?
First, Fortran is a statically and strongly typed language, making it
easier to write correct programs. Second, it provides strong support for whole-array
arithmetic, making it suitable to concisely express arithmetic operations over
large, multidimensional arrays. Third, Fortran is a natively parallel
programming language, with syntax that allows expressing parallel patterns
independent from the underlying hardware architecture, shared or distributed
memory alike. The most recent iteration of the Fortran standard is Fortran 2018
\citep{reid18}, which brought improvements to interoperability with C,
advanced parallel programming features such as teams and events, and
miscellaneous additions to the language and the standard library.
These features, as well as the continued development by
the Fortran Standards committee, provide a solid foundation for a
Fortran framework for neural networks and deep learning.

Development of this framework was largely motivated by my own desire to learn
how neural networks work inside out. In the process, I learned that there
may be a multifold usefulness to such library. First, it allows existing
Fortran developers without machine learning experience to ease into the field
without friction that comes up when one needs to learn both a new programming
language and a new concept.
Second, it allows existing machine learning practitioners who are not
familiar with modern Fortran to learn more about the language and its unique
capabilities, such as whole-array arithmetic and native distributed parallelism.
Finally, this framework can be used by scientists and engineers
to integrate machine learning flows into existing Fortran software, such as
those for civil engineering \citep{fischer08},
computational chemistry \citep{valiev10},
and numerical weather \citep{powers17}, ocean \citep{chassignet06},
wave \citep{donelan12}, and climate \citep{hurell13} prediction.
This paper thus serves as a gentle introduction to each of these audiences.

This paper describes the implementation of neural-fortran. While still only a
proof of concept, I demonstrate that its ease of use, serial performance, and
parallel scalability make it a viable candidate for use in production on its
own, or in integration with existing Fortran software. For brevity, I will not
go into much detail on how neural networks work in a mathematical sense, and I
will focus on the implementation details in the context of Fortran programming.
I first describe the features in Section \ref{section_features}, then I describe
the implementation in Section \ref{section_implementation}. I demonstrate the
use of the framework with a real-life example in Section \ref{section_examples}.
Then, I discuss the computational performance of the framework in Section
\ref{section_performance}. Concluding remarks are given in Section
\ref{section_conclusions}.

\section{Features} \label{section_features}

neural-fortran provides out-of-the-box support for:

\begin{itemize}
  \item Feed-forward neural networks with arbitrary number of hidden layers with
  arbitrary number of neurons;
  \item Several activation functions, including Gaussian, RELU, sigmoid, step,
  and tangent hyperbolic functions;
  \item Optimization using stochastic gradient descent (SGD) \citep{rumelhart86}
  and a quadratic cost function;
  \item Data-based parallelism using collective sum and broadcast operations;
  \item Single (\lstinline{real32}), double (\lstinline{real64}), or quadruple
  (\lstinline{real128}) precision floats, selected at compile time;
  \item Saving and loading networks to and from file.
\end{itemize}

Feed-forward neural networks are the simplest possible flavor of all neural
networks, and form a basis for other network flavors such as convolutional
\citep{krizhevsky12} and recurrent \citep{hochreiter97} neural networks.
Optimization algorithm and the choice of cost function allow a network
to learn from the data. Parallelism allows training or evaluating the
network using many processors at once.
Finally, the choice of single, double, or quadruple floating-point
precision is made available via type kind constants from the 
standard \lstinline{iso_fortran_env} module, and selected at 
compile-time using a preprocessor macro. 
While quadruple floating-point precision is likely an overkill
for most machine learning applications, it's trivially made 
available by most recent Fortran compilers, and may prove to 
be useful if an application that requires it comes about.

\section{Implementation} \label{section_implementation}

This section describes the fundamental data structures, networks and layers,
and the implementation of methods to compute the output of the network
(forward propagation), as well as learning from data (back-propagation).
Finally, we will go over the method to parallelize the training of the network.

\subsection{Networks, layers, and neurons}

At a high level, a neural network is made of an arbitrary number of layers,
each having an arbitrary number of neurons. Each neuron has at least two
fundamental properties: A bias $b$, which determines how easy it is to activate
the neuron, and a weight $w$ for every connection to other neurons, determining
the strenght of the connection. A feed-forward neural network is also often
called dense because each neuron in one layer is connected to all neurons in
the next layer. Due to this unstructured nature, a contiguous Fortran array is
not a convenient choice to model a network. Instead, we can model a network
with a derived type, one of its components being a dynamic array of layers
(Listing \ref{network_listing}).

\begin{minipage}{\linewidth}
\begin{lstlisting}[caption={Definition of the network class. Type-bound methods are omitted for brevity.}, captionpos=b, label={network_listing}]
type :: network_type
  type(layer_type), allocatable :: layers(:)
  integer, allocatable :: dims(:)
  procedure(activation_function), &
    pointer, nopass :: activation => null()
  procedure(activation_function), &
    pointer, nopass :: activation_prime => null()
contains
  ...
end type network_type
\end{lstlisting}
\end{minipage}

The key component in the network class is the allocatable array of layers
instances. The \lstinline{allocatable} attribute is essential to allowing the
size and structure of the network to be determined at run time rather than compile time.
We keep track of two procedure pointers at this point, one for the activation
function and another for the derivative of the activation function. The former
is input by the user, or given a default value if not specified. The latter is
determined by the activation function. While not crucial for functionality,
we keep the component \lstinline{dims} for housekeeping.
The network class has several private and public methods, which are omitted in
Listing \ref{network_listing}. We will look into these methods in more detail
soon.

To create and initialize a network instance from a single line, we need to
build all the set up logic into a custom constructor function (Listing
\ref{network_constructor_listing}).

\begin{minipage}{\linewidth}
\begin{lstlisting}[caption={Custom network constructor.}, captionpos=b, label={network_constructor_listing}]
type(network_type) function &
  net_constructor(dims, activation) result(net)
  integer(ik), intent(in) :: dims(:)
  character(len=*), intent(in), optional :: &
    activation
  call net % init(dims)
  if (present(activation)) then
    call net % set_activation(activation)
  else
    call net % set_activation('sigmoid')
  end if
  call net % sync(1)
end function net_constructor
\end{lstlisting}
\end{minipage}

The network constructor accepts two input arguments. First is a rank-1 array of
integers \lstinline{dims}. This array controls how many layers to allocate in
the network, and how many neurons should each layer have. \lstinline{size(dims)}
is the total number of layers, including input and output layers. The second
argument is optional, and it is a character string that contains the name of the
activation function to use in the network. If not present, we default to a
sigmoid function, $\sigma(x) = 1 / \left( 1 + e^{-x} \right)$,
a commonly used function for neural network applications.
The first step that we take in the constructor is to call the \lstinline{init()} method,
which goes on to allocate the layers and their components in memory, and to
initialize their values. Second, we set the procedure pointers for the activation
function and its derivative. Finally, the \lstinline{sync()} method broadcasts
the state of the network to all other processors if executed in parallel.

With our custom constructor defined, a network instance can be created from the
client code as shown in Listing \ref{using_constructor_listing}.

\begin{minipage}{\linewidth}
\begin{lstlisting}[caption={Creating a network instance in the client code.}, captionpos=b, label={using_constructor_listing}]
use mod_network, only: network_type
type(network_type) :: net
net = network_type([3, 5, 2], 'tanh')
\end{lstlisting}
\end{minipage}

In this example, we created a network instance with an input layer of 3 neurons,
one hidden layer of 5 neurons, an output layer with 2 neurons, and a tangent
hyperbolic activation function. Recall that providing the name of the activation
function is optional. If omitted, it will default to a sigmoid function.

Just as a network can be made of any number of layers, so can a layer be made
of any number of neurons. For the basic feed-forward network architecture,
we don't need special treatment of neurons using derived types, and modeling
their properties as Fortran arrays inside the layer class is straightforward
and sufficient (Listing \ref{layer_listing}).

\begin{minipage}{\linewidth}
\begin{lstlisting}[caption={Definition of the layer class.}, captionpos=b, label={layer_listing}]
type :: layer_type
  real(rk), allocatable :: a(:) ! activations
  real(rk), allocatable :: b(:) ! biases
  real(rk), allocatable :: w(:,:) ! weights
  real(rk), allocatable :: z(:) ! temp. array
end type layer_type
\end{lstlisting}
\end{minipage}

The layer is thus simply made of contiguous allocatable arrays for activations,
biases, weights, and a temporary array to be used in the backpropagation step.
Note that unlike the activations and biases, the weights are of rank 2, one for
each neuron in this layer, and the other for all the neurons in the next layer.
The kind constant \lstinline{rk} refers to the real kind, and can be selected
by the user at compile time as either \lstinline{real32} (single precision,
default), \lstinline{real64} (double precision), and \lstinline{real128}
(quadruple precision, if available).

Before we begin any training, it is important to initialize the network with
quasi-random weights and biases to make it as little biased as possible.
To that end, we initialize the weights using a simplified variant of Xavier's
initialization \citep{glorot10}, as shown in Listing \ref{layer_constructor_listing}.

\begin{minipage}{\linewidth}
\begin{lstlisting}[caption={Layer constructor.}, captionpos=b, label={layer_constructor_listing}]
type(layer_type) function &
  constructor(this_size, next_size) result(layer)
  integer(ik), intent(in) :: this_size, next_size
  allocate(layer % a(this_size))
  allocate(layer % z(this_size))
  layer % a = 0
  layer % z = 0
  layer % w = randn(this_size, next_size) &
             / this_size
  layer % b = randn(this_size)
end function constructor
\end{lstlisting}
\end{minipage}

All weights are initialized as random numbers with a normal distribution
centered on zero, and normalized by the number of neurons in the layer. The
goal of normalization is to prevent the saturation of the network output
for large networks.
Activations are computed and stored during the evaluation of the network output,
so they are initialized to zero.

Note that neural-fortran at this time does not treat individual neurons
with their own data structure. While this approach allows for simpler
implementation, it does impose limitations on what kinds of neural networks
can be built. Specifically, the current design allows for a family of classic,
feed-forward neural networks. More complex network architectures, such
as recurrent neural networks (RNN, \cite{hochreiter97}) which are well suited
for natural language processing, would require additional logic either by
modeling individual neurons with derived types, or using Fortran pointers.
Efforts to generalize the possible network architectures that can be constructed
in neural-fortran will be considered in the future.

\subsection{Computing the network output}

Given a state of a network's weights and biases, and a sample of input data,
we can compute the output of the network using the so-called forward propagation.
This procedure involves computing the activations of all neurons layer by layer,
starting from the input layer and finishing with the output layer. The
activations of each previous layer take the role of input data.

The output of a neuron is calculated as the weighted sum of neuron outputs
from the previous layer, plus the bias of that neuron, result of which is passed
to an activation function. For each neuron $i$ in layer $j$, we evaluate its
output as:

\begin{equation}
  \label{eq:activation}
  a_{ij} = \sigma \left( \sum_i w_{i(j-1)} x_{i(j-1)} + b_{ij} \right)
\end{equation}

where $\sigma$ is the activation function.
$\sum w_{i(j-1)} x_{i(j-1)}$ is an expanded form of a dot product,
so we can write this more concisely in vector form as:

\begin{equation}
  \label{eq:activation_vector}
  \mathbf{a}_j = \sigma \left( \mathbf{w}_{j-1} \cdot \mathbf{x}_{j-1} + \mathbf{b}_j \right)
\end{equation}

Fortran's whole array arithmetic allows us to express Eq. \ref{eq:activation_vector}
in code almost identically, as shown in Listing \ref{fwdprop_listing}.

\begin{minipage}{\linewidth}
\begin{lstlisting}[caption={A subroutine to perform the forward propagation of the network, and store intermediate activations for later use.}, captionpos=b, label={fwdprop_listing}]
pure subroutine fwdprop(self, x)
  class(network_type), intent(in out) :: self
  real(rk), intent(in) :: x(:)
  integer(ik) :: n
  associate(layers => self % layers)
    layers(1) % a = x
    do n = 2, size(layers)
      layers(n) % z = &
        matmul(transpose(layers(n-1) % w), &
        layers(n-1) % a) + layers(n) % b
      layers(n) % a = &
        self % activation(layers(n) % z)
    end do
  end associate
end subroutine fwdprop
\end{lstlisting}
\end{minipage}

The activations for the first (input) layer are just input data $x$. The actual
computation of activations is done from the second layer and onward. For each
following layer, the input data are the activations of the previous layer.
For the evaluation of $\mathbf{w}_{j-1} \cdot \mathbf{x}_{j-1}$, we use
the intrinsic \lstinline{matmul} rather than the \lstinline{dot_product}
because \lstinline{layer_type 
in this layer, and one rank for all neurons in the following (connecting) layer.
The \lstinline{associate} construct is here only to make the code
less verbose, and is otherwise unnecessary for the implementation.

While the \lstinline{fwdprop} subroutine is declared as pure, it does modify
the state of the network, by storing $\mathbf{w}_{j-1} \cdot \mathbf{x}_{j-1} + \mathbf{b}_j$
of each layer as it propagates through the layers. Storing these values is
necessary for the back-propagation algorithm used for updating the weights and
biases (see subsection \ref{backprop}).

A variant of \lstinline{fwdprop} that doesn't store any intermediate values
but only returns the output of the network is also available as a pure function
\lstinline{network_type 
of training of the network, such as testing or prediction.

\subsection{Optimization} \label{backprop}

The optimization method used in neural-fortran is the backpropagation with the
stochastic gradient descent \citep{rumelhart86}. I omit the details for brevity,
but the essence of the method can be described in three steps:

\begin{enumerate}
  \item Evaluate the cost function $C$ of the network, that is, calculate the error
  of the network output relative to the training or testing data;
  \item Find the gradient of $C$ in the weight-bias space: $\left( \dfrac{\partial C}{\partial w}, \dfrac{\partial C}{\partial b} \right)$. The gradient vector informs us about
  the direction in which $C$ decreases the fastest;
  \item Update weights and biases in all layers such that $C$ decreases.
\end{enumerate}

Thus the name gradient descent. The second step is the core of the method.
The "stochastic" part comes about in the way that the input data is 
shuffled during the training.

Back to our code. Assuming we have done a forward propagation step and updated
the activations in all layers, a backward propagation step will compute
the weight and bias tendencies \lstinline{dw} and \lstinline{db}, as shown in
Listing \ref{backprop_listing}.

\begin{lstlisting}[caption={A subroutine to perform the backpropagation using gradient descent, and return the weight and bias tendencies.}, captionpos=b, float=*h, label={backprop_listing}]
pure subroutine backprop(self, y, dw, db)
  class(network_type), intent(in out) :: self
  real(rk), intent(in) :: y(:)
  type(array2d), allocatable, intent(out) :: dw(:)
  type(array1d), allocatable, intent(out) :: db(:)
  integer :: n, nm

  associate(dims => self % dims, layers => self % layers)

    call db_init(db, dims)
    call dw_init(dw, dims)

    n = size(dims)
    db(n) % array = (layers(n) % a - y) * self % activation_prime(layers(n) % z)
    dw(n-1) % array = matmul(reshape(layers(n-1) % a, [dims(n-1), 1]), &
                                reshape(db(n) % array, [1, dims(n)]))

    do n = size(dims) - 1, 2, -1
      db(n) % array = matmul(layers(n) % w, db(n+1) % array) &
                     * self % activation_prime(layers(n) % z)
      dw(n-1) % array = matmul(reshape(layers(n-1) % a, [dims(n-1), 1]), &
                                  reshape(db(n) % array, [1, dims(n)]))
    end do

  end associate

end subroutine backprop
\end{lstlisting}

Unlike the forward propagation step, backward propagation does not modify
the state of the network, but returns the weight and bias tendencies that
would minimize the cost function given data \lstinline{y}. Updating the weights
and biases is just a matter of iterating over all the layers and applying the
tendencies returned by \lstinline{network_type 
method \lstinline{network_type 
is omitted in this paper for brevity.

\subsection{The main training loop}

Now that we have the essential building blocks of a neural network's training
loop (forward propagation, backward propagation, and update), we can put them
together in a sequence to make a convenience high-level training procedure.

The subroutine \lstinline{train_single} takes a single set of input data x and y,
and the learning rate \lstinline{eta}, and updates its weights and biases
accordingly (Listing \ref{train_loop_listing}).

\begin{minipage}{\linewidth}
\begin{lstlisting}[caption={A training procedure for a single data sample.}, captionpos=b, label={train_loop_listing}]
pure subroutine train_single(self, x, y, eta)
  class(network_type), intent(in out) :: self
  real(rk), intent(in) :: x(:), y(:), eta
  type(array2d), allocatable :: dw(:)
  type(array1d), allocatable :: db(:)
  call self % fwdprop(x)
  call self % backprop(y, dw, db)
  call self % update(dw, db, eta)
end subroutine train_single
\end{lstlisting}
\end{minipage}

This method takes a sample of input data $x$, output data $y$, and a learning
rate as input arguments. We first forward propagate the network using the input
data. At this stage, all activations in the network are updated and stored.
Second, we perform the backpropagation using the output data. At this step,
we get weight and bias tendencies. Finally, we pass the tendencies and the
learning rate to the update method which increments the state of the network.

More commonly than not, bias and weight tendencies are computed and accumulated
over a long sequence of inputs and outputs (a batch), and applied
at the very end. Thus we also need the \lstinline{train_batch} subroutine
which accepts rank-2 arrays $x$ and $y$ as input arguments, first dimension
corresponding to the shapes of input and output layers, respectively, and the
second dimension corresponding to the size of the data batch
(Listing \ref{train_batch_listing}).

\begin{minipage}{\linewidth}
\begin{lstlisting}[caption={A variant of the training method that accepts batches of data.}, captionpos=b, label={train_batch_listing}]
subroutine train_batch(self, x, y, eta)
  class(network_type), intent(in out) :: self
  real(rk), intent(in) :: x(:,:), y(:,:), eta
  ...
end subroutine train_batch
\end{lstlisting}
\end{minipage}

Finally, access to either variant is made via the generic procedure
\lstinline{train} (Listing \ref{train_generic_listing}).

\begin{minipage}{\linewidth}
\begin{lstlisting}[caption={Overloading specific training procedures with a generic name.}, captionpos=b, label={train_generic_listing}]
generic, public :: &
  train => train_batch, train_single
\end{lstlisting}
\end{minipage}

The user can thus perform training using a single sample of data,
or a batch of data, using the same generic procedure. The correct
specific procedure is determined by the compiler depending on the rank of
input data. Example:

\begin{minipage}{\linewidth}
\begin{lstlisting}[caption={Using the generic training procedure with single samples and batches of data.}, captionpos=b, label={train_generic_usage_listing}]
! invokes network % train_single()
network % train(x(:,n), y(:,n), eta)

! invokes network % train_batch()
network % train(x(:,:), y(:,:), eta)
\end{lstlisting}
\end{minipage}

This subsection completes the training sequence. The last remaining piece
is the implementation of the parallelism, described in the following section.

\subsection{Parallelism}

There are two main approaches to parallelizing neural networks:
Data-based and model-based parallelism.

In the data-based parallelism approach, a training data batch is evenly distributed
between the parallel processors, which can be working in either
shared or distributed memory. Each processor calculates the weight and
bias tendencies based on their piece of the data batch, and a collective
sum operation is done over all parallel processes to calculate the
total update to the weights and biases of the network.
It is crucial that the randomly initialized weights and
biases are equal on all parallel processors.
This can be done either by all processes using the same random seed
at initialization, or by broadcasing weights and biases from one process to all
others. neural-fortran adopts the latter approach.

The data-based parallelism in neural-fortran is accomplished with the
following steps:

\begin{enumerate}

  \item Create the network instance on each image by invoking the
  \lstinline{network_type} constructor. Inside the constructor, the networks
  are synchronized by broadcasting the initial weights and biases from the
  image 1 to all other images. This is done using the intrinsic
  \lstinline{co_broadcast} subroutine. As the synchonization is built into the
  type constructor (see Listing \ref{network_constructor_listing}), the user
  simply has to create the network instance as if they were working in serial mode.

  \item Each parallel process calculates its own weight and bias tendencies given
  their piece of the training data batch.

  \item Finally, the weight and bias tendencies are summed up across all parallel
  processes using the collective sum method:

  \begin{minipage}{\linewidth}
  \begin{lstlisting}
  if (num_images() > 1) then
    call dw_co_sum(dw_batch)
    call db_co_sum(db_batch)
  end if
  \end{lstlisting}
  \end{minipage}

  Once the collective sum operation is complete, each process updates their
  own copy of the network. \lstinline{dw_co_sum} and \lstinline{db_co_sum} are
  thin wrappers around \lstinline{co_sum} for arrays of derived types that
  hold the weight and bias tendencies, respectively.

\end{enumerate}

neural-fortran thus relies exclusively on collective subroutines
\lstinline{co_sum} (for the collective sum of weight and bias tendencies) and
\lstinline{co_broadcast} (for broadcasting randomly initialized weights and biases
across all processes), introduced in Fortran 2018 \citep{reid18}.
Collective subroutines let the programmer express a family of parallel
algorithms without involving coarrays for explicit data communication.
Being easy to write, debug, and reason about, they are an extremely valuable
new feature of the language.

In contrast to the data-based parallelism, the model-based parallelism entails
parallelizing the dot product and matrix multiplication operations used in
the forward and backward propagation steps of the training procedure.
Model-based parallelism is currently not implemented in neural-fortran.
The first step toward such capability would likely involve interfacing
external libraries that provide parallel implementations of
\lstinline{dot_product} and \lstinline{matmul}, such as the Intel Math Kernel
Library, OpenBLAS (\url{http://www.openblas.net}), or ATLAS \citep{whaley01, whaley04}.
As as the \lstinline{matmul} invocation in \lstinline{fwdprop} and
\lstinline{backprop} would be automatically replaced by the compiler with the
optimized and parallel procedure provided by the linked library, model-based
parallelism could thus be accomplished without any further modification to
the code.

The data- and model-based parallelism approaches are thus decoupled and
independent from each other. For large systems with distributed nodes with
multiple cores on each, a hybrid approach could be adopted: Intra-node (shared memory)
parallelization of \lstinline{matmul} via external linear algebra library,
and inter-node (distributed memory) parallelization via Fortran 2018 collective
subroutines. Viability and performance of such hybrid approach will be explored
in the future and is beyond the scope of this paper.

\section{Recognizing handwritten digits} \label{section_examples}

To demonstrate the use of neural-fortran we will look at an example of training
a network to recognize handwritten digits using the MNIST database of images
\citep{lecun98} (Figure \ref{fig_mnist_examples}). MNIST is commonly used in
teaching machine learning methods, and for development and testing of new algorithms.
This dataset is included in the neural-fortran repository as is a good example
for beginners to get started and play with the framework.
A convenience Fortran subroutine to load the dataset
is available in the \lstinline{mod_io} module.

\begin{figure}[H]
  \centering
  \includegraphics[width=\columnwidth]{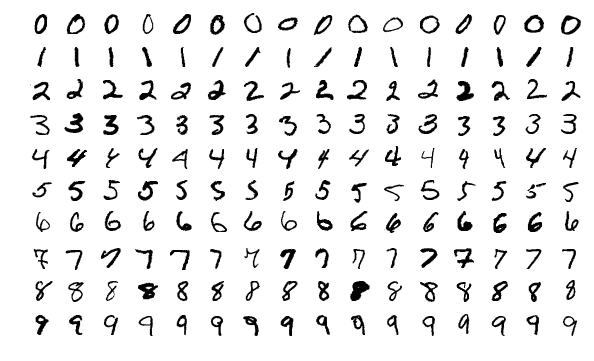}
  \caption{A small set of data samples from the MNIST database.}
  \label{fig_mnist_examples}
\end{figure}

The MNIST dataset consists of 70000 images of handwritten digits, 
and the same number of labels denoting the value in each image. 
A label is a qualitative value that classifies the input image.
In this paper, 50000 images will be used for training, and 10000 for validation.
Each image is a 28 by 28 pixel greyscale scan of a numerical digit in the range [0-9]. 
The labels for this dataset are thus numerical values in the same range. 
Figure \ref{fig_mnist_example_digit} offers a close-up view of one such image,
labeled as number 2.

\begin{figure}[H]
  \centering
  \includegraphics[width=\columnwidth]{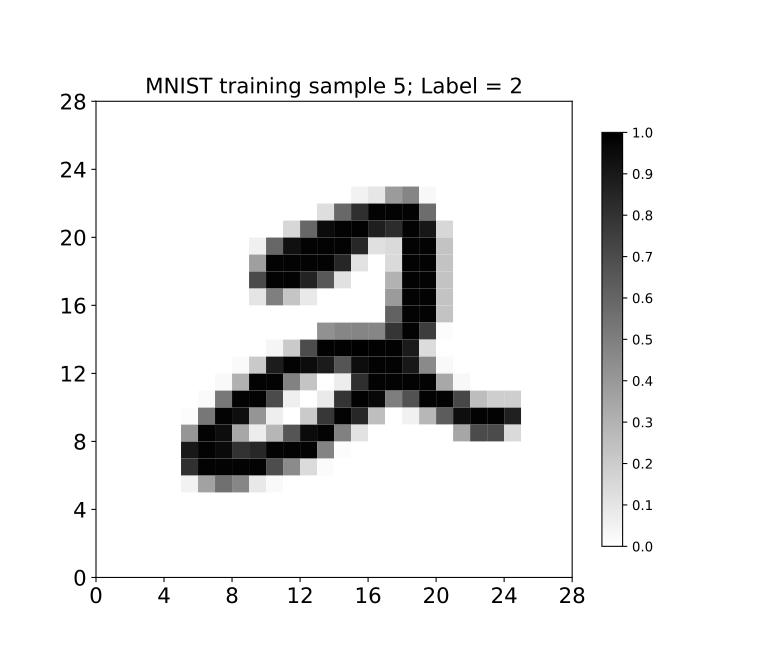}
  \caption{A close-up view at a single digit sample of 28 by 28 pixels.
           The value of each pixel is the greyscale intensity, zero for white
           and one for black.}
  \label{fig_mnist_example_digit}
\end{figure}

In this example, we will load the MNIST dataset into memory, the data from
each image being represented as a rank-1 real array of 784 elements, with
magnitude in the range [0-1]. The pixel values from each image will serve as the
input data $x$ to our network. The labels will serve as the output data $y$.
Rather than a single number indicating the value of the label, we will use
an array of 10 elements, one for each digit. The values of the elements will
be zero for all elements except the one corresponding to the value of the label,
which will take the value of one.
Listing \ref{mnist_example_listing} shows the complete program.

\begin{lstlisting}[caption={The complete program to train and evaluate a neural network using the MNIST dataset of handwritten digits.}, captionpos=b, float=*, label={mnist_example_listing}]
program example_mnist

  use mod_kinds, only: ik, rk
  use mod_mnist, only: label_digits, load_mnist
  use mod_network, only: network_type

  implicit none

  real(rk), allocatable :: tr_images(:,:), tr_labels(:)
  real(rk), allocatable :: te_images(:,:), te_labels(:)
  real(rk), allocatable :: input(:,:), output(:,:)

  type(network_type) :: net

  integer(ik) :: i, n, num_epochs
  integer(ik) :: batch_size, batch_start, batch_end
  real(rk) :: pos

  call load_mnist(tr_images, tr_labels, te_images, te_labels)

  net = network_type([784, 30, 10])

  batch_size = 1000
  num_epochs = 30

  if (this_image() == 1) then
    write(*, '(a,f5.2,a)') 'Initial accuracy: ',&
      net % accuracy(te_images, label_digits(te_labels)) * 100, ' %'
  end if

  epochs: do n = 1, num_epochs
    mini_batches: do i = 1, size(tr_labels) / batch_size

      ! pull a random mini-batch from the dataset
      call random_number(pos)
      batch_start = int(pos * (size(tr_labels) - batch_size + 1))
      batch_end = batch_start + batch_size - 1

      ! prepare mini-batch
      input = tr_images(:,batch_start:batch_end)
      output = label_digits(tr_labels(batch_start:batch_end))

      ! train the network on the mini-batch
      call net % train(input, output, eta=3._rk)

    end do mini_batches

    if (this_image() == 1) then
      write(*, '(a,i2,a,f5.2,a)') 'Epoch ', n, ' done, Accuracy: ',&
        net % accuracy(te_images, label_digits(te_labels)) * 100, ' %'
    end if

  end do epochs

end program example_mnist
\end{lstlisting}

This program has 3 key pieces. First, it loads the training and testing data
into the memory by calling the \lstinline{load_mnist} subroutine. This subroutine
returns input and output data arrays that are in the appropriate shape for
working with neural-fortran. While not obvious from the listing, the training
dataset consists of 50000 images, and the testing dataset consists of 10000
images. The distinction between the two is built into the data loader.
It is important for the validation dataset to be completely distinct 
from the training dataset to ensure that the model can learn in a general way. 

Second, we create our network instance, in this case a network with one input
layer with 784 neurons, one hidden layer with 30 neurons, and one output layer
with 10 neurons. Recall that the sizes of input and output layers are not
arbitrary. The input layer size must match the size of the input data sample.
In case of MNIST, each image is 28 by 28 pixels, a total of 784 pixels.
Similarly, the size of the output layer must match the size of the output
data sample, an array of 10 elements, one for each digit.
The rationale for transforming the value of the image label into an array
of 10 binary values is empirical. 
As the weights and biases of the network are adjusted in the training 
process, the network assigns higher probability to one or more digits.
The output of the network is thus the probability that the input image represents
each of the 10 digits. For a well trained network and a clean input image, 
the network should output one value that is approximately 1, and nine others
that are approximately 0. This binary label representation thus allows for 
probabilistic output of the network. While other output layer configurations 
are possible, they don't tend to yield as good results.

In the third step, we begin the training loop, which is performed for a set
number of epochs. An epoch refers to one round of cycling through the whole
training dataset, in our case, 50000 images. Within each epoch, however,
we iterate over a number of so-called mini-batches, which are the subsets of
the whole dataset. The number of mini-batches is controlled by the batch size,
in our case 1000, meaning that a total of 50 training iterations will be done
in each epoch. The random choice of the starting index of the mini-batch
makes this gradient descent approach a stochastic one. 
Randomly sampling the training data in each batch prevents the network
from converging toward a state that is specific to the ordering of the data.
Note that in this simplistic implementation using the \lstinline{random_number} subroutine, 
not all data samples will be used even for a large number of epochs,
and there will be some overlap within each epoch. 
While this works well enough for this example, 
more sophisticated shuffling should be used in production.

As we saw in Section \ref{backprop}, the learning rate \lstinline{eta} is the
multiplier to weight and bias tendencies at the time of the network update.
Its value is somewhat arbitrary and varies between applications. An optimal
value of \lstinline{eta} thus needs to be determined by experimentation whenever
we work with new datasets or network structures. A value of \lstinline{eta}
that is too high may lead to never converging to a local minimum of the cost
function. On the other hand, a value too low may lead to a slow and
computationally expensive training procedure.

The neural-fortran API used in this example is rather simple. We used only
a network constructor and two methods, \lstinline{accuracy()}, and
\lstinline{train()}. All other code was for the purpose of loading and
preparing the training data, slicing mini-batches out of the whole dataset,
and other housekeeping.

Running this program outputs the following:

\begin{minipage}{\linewidth}
\begin{lstlisting}[caption={Output from the MNIST training example program. Some lines omitted for brevity.}, captionpos=b]
Initial accuracy: 10.09 %
Epoch  1 done, Accuracy: 27.91 %
Epoch  2 done, Accuracy: 53.17 %
Epoch  3 done, Accuracy: 75.16 %
Epoch  4 done, Accuracy: 82.96 %
Epoch  5 done, Accuracy: 87.24 %
...
Epoch 30 done, Accuracy: 93.39 %
\end{lstlisting}
\end{minipage}

The program first outputs the initial accuracy, which is approximately 10\%.
The accuracy is calculated as the number of digits (out of 10000, the size
of the testing dataset) that are correctly predicted by the network.
Since we have not done any training at this point, the accuracy is equivalent
to that of a random guess of one in ten digits. With each epoch, the accuracy
steadily improves and exceeds 93\% at the 30th epoch. While this is not an
extraordinarily high accuracy for handwritten digit recognition, it demonstrates
that our simple little network works as intended. For example, a deep 6-layer
network with several thousand neurons was shown to achieve accuracy of
99.65\% \citep{ciresan10}.

\begin{figure}[H]
  \centering
  \includegraphics[width=\columnwidth]{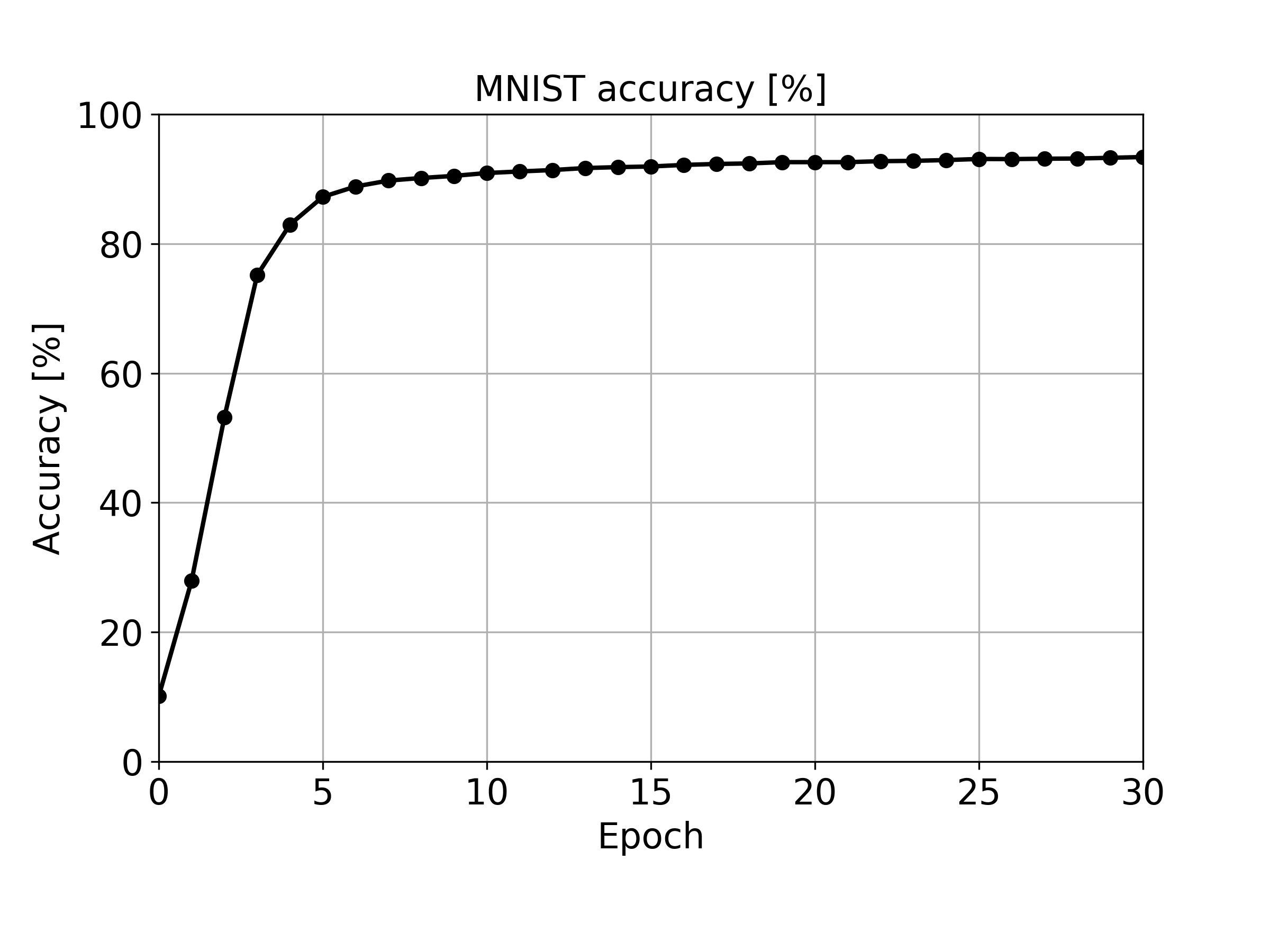}
  \caption{Accuracy as function of the number of epochs in the MNIST example.}
  \label{fig_mnist_accuracy}
\end{figure}

The network accuracy as function of number of epochs is shown on Figure
\ref{fig_mnist_accuracy}. It shows that the fastest learning occurs during
the first five epochs, beyond which the network learns slowly and eventually
plateaus.
The rate of the increase in accuracy, and the final value at which the accuracy
asymptotes is sensitive to almost every aspect of the network:
Number of layers and neurons in each layer, batch size,
learning rate, the choice of activation and cost functions, the initialization
method for weights and biases, and more.

\section{Computational performance} \label{section_performance}

We evaluate two aspects of computational performance of neural-fortran:
Serial performance in comparison with an established machine learning framework,
and parallel scalability. Both benchmarks were done with the following
configuration:

\begin{itemize}
  \item Processor: Intel Xeon Platinum 8168 (2.70GHz)
  \item Operating System: Fedora 28 64-bit
  \item Compiler: GNU Fortran 8.2.1
  \item Libraries:
  \begin{itemize}
    \item OpenMPI 2.1.1
    \item OpenCoarrays 2.3.1 \citep{fanfarillo14}
  \end{itemize}
\end{itemize}

\subsection{Serial performance}

To evaluate the serial (single-core) performance of neural-fortran,
we compare the elapsed training time of neural-fortran against an established
and mature machine learning framework, Keras 2.2.4 \citep{chollet15} using
Tensorflow 1.12.0 \citep{abadi16} as the computational backend. This framework
was configured in a Python 3.6.8 environment with numpy 1.16.1 and scipy 1.2.1
on the same system used to run neural-fortran.

The performance is compared using the handwritten digits example
with MNIST image dataset, with the same layer configuration as that used
in Section \ref{section_examples} (784-30-10 neurons with sigmoid activations).
The Keras network was configured in the exact same fashion, with stochastic
gradient descent optimizer and the quadratic cost function. The default batch
size for Keras is 32 if not specified, so we use that value in this comparison.
A total of 10 training epochs are done in each run.
As the Tensorflow backend engages multithreading by default, we explicitly
limit the number of cores and threads to one, for fairness.
A total of 5 runs are made, the mean and standard deviation of which are shown
in Table \ref{serial_table}.

\begin{center}
\begin{table}[h]
\caption{Computational performance of neural-fortran and Keras+Tensorflow
         in the MNIST training example. Elapsed time shown is the mean $\pm$
         standard deviation of 5 repeated runs.}
\label{serial_table}
\begin{tabular}{c|c|c}
  \hline
  Framework & Elapsed (s) & Memory use (MB) \\
  \hline
  neural-fortran     & 13.933 $\pm$ 0.378 & 220 \\
  Keras + Tensorflow & 12.419 $\pm$ 0.474 & 359 \\
  \hline
\end{tabular}
\end{table}
\end{center}

This basic comparison of serial performance between neural-fortran and
Keras + Tensorflow in the MNIST training example demonstrates that they are
similar to each other. The elapsed time is 12\% longer in the case
of neural-fortran on average, and 20\% less variable between individual runs.
This is likely because the neural-fortran program is more "bare-bones", in
contrast to the Keras program that at a high level goes through the Python
interpreter, which may add some unpredictable overhead. neural-fortran also
uses about 39\% less memory than Keras with Tensorflow, meaning that it could
be used to solve larger problems within equivalent memory constraints. Although
Keras + Tensorflow marginally outperforms neural-fortran in terms of average
run-time, it's important to note that neural-fortran's core computations are
not optimized, but expressed as a a concise proof of concept.
Considering that Keras and Tensorflow are mature frameworks that are used
widely in production and at scale, this suggests that neural-fortran has
potential to be computationally competitive in production, especially if
optimizations are made to the linear algebra operations.

Finally, it is worth comparing the ease of use of neural-fortran to that of Keras. 
The most straightforward metric is the number of lines of code needed to
construct, train, and evaluate the network. The neural-fortran program used
in this benchmark consists of 35 lines of code excluding comments,
whereas the Keras script consists of 41 lines of code. Five of these lines
were needed to constrain the Keras network to use a single thread.
The amount of code needed for this example is thus almost equivalent
between the two frameworks.
The Python script used to construct, train, and time the Keras network,
and the neural-fortran program used in this benchmark are available in the
source code repository for this paper.

\subsection{Parallel scaling}

With a basic measure of serial performance established, let's evaluate
neural-fortran's parallel scaling. All network parameters are the same as in
the serial case, except for the batch size, which is now set to 1200.
A large batch size is important for parallel scaling because a single batch is
distributed evenly between the parallel processes.
Unlike in the serial benchmarks, here we time only the traning portion
of the code and exclude the loading of the dataset. Elapsed times on up to
12 parallel images on a shared-memory system are shown in Figure
\ref{fig_mnist_elapsed}. The elapsed time decreases monotonically from over
12 s in serial mode to under 2 s on 12 parallel cores.

\begin{figure}[H]
  \centering
  \includegraphics[width=\columnwidth]{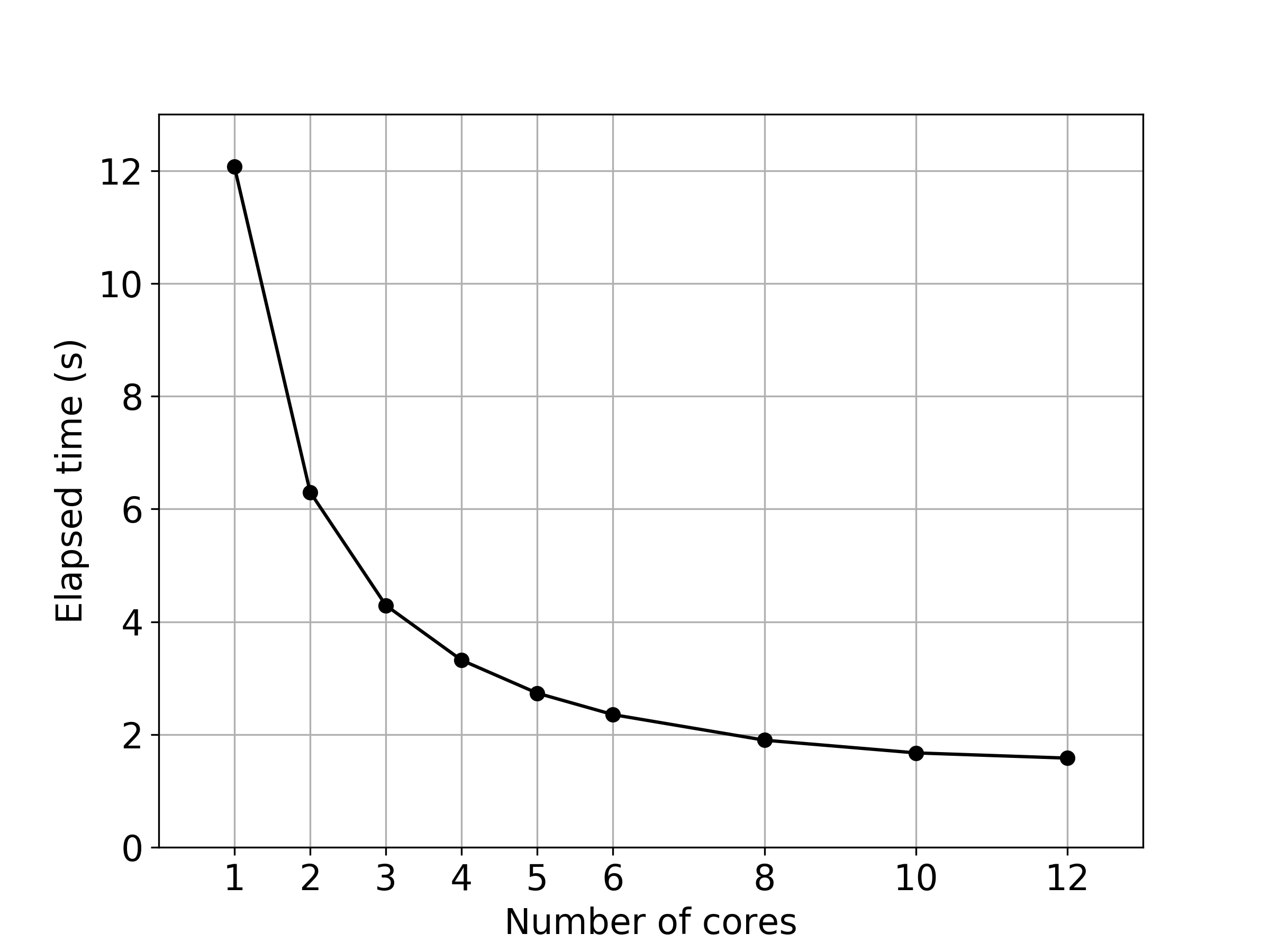}
  \caption{Mean elapsed time as function of number of cores in the MNIST training example.}
  \label{fig_mnist_elapsed}
\end{figure}

Next we look at the parallel efficiency (PE), which is the serial elapsed time
divided by the parallel elapsed time and the number of processes,
$PE = t(1) / (n t(n))$ (Figure \ref{fig_mnist_efficiency}).
On the high end, perfect scaling corresponds $PE = 1$.
On the low end, no parallel speed-up relative to the serial performance,
meaning $t(n) = t(1)$ for any value of $n$, corresponds to $PE = 1/n$ and 
is shown using the dashed line.

\begin{figure}[H]
  \centering
  \includegraphics[width=\columnwidth]{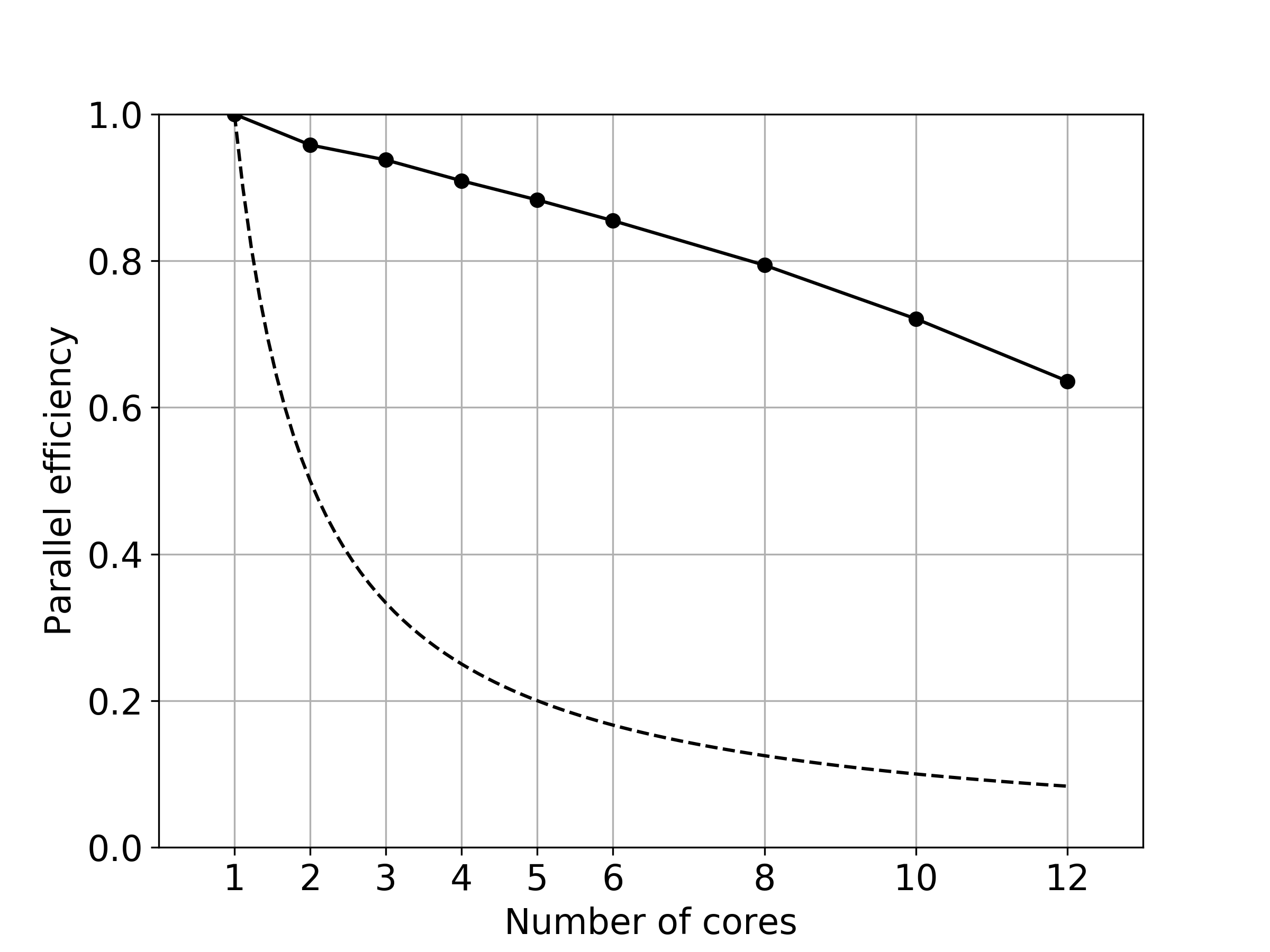}
	\caption{Parallel efficiency as function of number of cores in the MNIST training example (solid line),
	   and zero speed-up relative to the serial mode, $PE = 1/n$ (dashed line).}
  \label{fig_mnist_efficiency}
\end{figure}

The mean elapsed times and their standard deviation (out of five runs) are
given in Table \ref{table_parallel}.

\begin{center}
\begin{table}[h]
\caption{Parallel scaling of neural-fortran in the MNIST training example.}
\label{table_parallel}
\begin{tabular}{c|c|c}
  \hline
  Cores & Elapsed (s) & Parallel efficiency \\
  \hline
  1 & 12.068 $\pm$ 0.136 & 1.000 \\
  2 & 6.298 $\pm$ 0.043 & 0.958 \\
  3 & 4.290 $\pm$ 0.026 & 0.938 \\
  4 & 3.318 $\pm$ 0.005 & 0.909 \\
  5 & 2.733 $\pm$ 0.005 & 0.883 \\
  6 & 2.353 $\pm$ 0.016 & 0.855 \\
  8 & 1.900 $\pm$ 0.019 & 0.794 \\
  10 & 1.674 $\pm$ 0.012 & 0.721 \\
  12 & 1.581 $\pm$ 0.046 & 0.636 \\
  \hline
\end{tabular}
\end{table}
\end{center}

Going from a single to 12 parallel cores, the parallel efficiency drops
to 64\%. With increasing number of parallel processes, the computational
load decreases because the batch of the same size is divided into smaller pieces.
Conversely, the amount of communication between processes that needs to be
done by the collective sum of weight and bias tendencies increases. This is
analogous to the scalability limits in fluid dynamics, except that here the
parallel operation is global, rather than only between the neighboring
processes. 
Nevertheless, the measured $PE$ remains well above the zero-speed-up line
for all configurations, indicating that in this example neural-fortran
can efficiently use all cores in a high-end shared memory system.
The ability to explicitly allocate any number
of cores in the system, or to execute the training on a distributed-memory
system without any change to the code, are unique features of neural-fortran
that are not trivial to accomplish in a framework like Keras or Tensorflow.

\section{Conclusions} \label{section_conclusions}

This paper described the implementation of neural-fortran, a parallel framework
for neural networks and deep learning. It allows constructing feed-forward
neural networks with any number of layers and neurons and offers several
activation functions and learning via stochastic gradient descent.
neural-fortran also leverages the collecive sum and broadcast procedures
introduced in Fortran 2018 to accomplish data-based parallelism.
Model-based parallelism can be enabled by linking to external linear algebra
libraries such the Intel MKL or ATLAS. Use of neural-fortran was demonstrated
in an example of training the network to recognize handwritten digits.
In a rudimentary benchmark, computational performance of neural-fortran is
comparable to an established and mature framework (Keras + Tensorflow),
making it a viable candidate for use in production.
While currently only a proof of concept with a limited number of features,
neural-fortran provides a good foundation for further development and
applications.

There are two potential benefits of using neural-fortran in new
software development and machine learning projects.
One is tight integration with existing Fortran code, such as computational
fluid dynamics for engineering applications, or numerical weather prediction.
For example, efforts to augment existing climate modeling capabilities are underway,
one instance being the Earth Machine project \citep{schneider17, voosen18}.
Such projects work with large Fortran code bases with many parametric and
empirical sub-models that could be largely accelerated via tight integration
with a machine learning framework such as neural-fortran.
Another benefit is the ease of use for experienced Fortran programmers who
want to get started with machine learning and neural networks without needing
to also learn another programming language or framework. For this reason, any
further development of neural-fortran will maintain ease of use as its primary
focus.

\begin{acks}
I am grateful to Arjen Markus and Izaak Beekman who reviewed and helped improve this article.
I am also thankful to the Sourcery Institute team for their work on OpenCoarrays. 
Michael Hirsch helped setting up the continuous integration for neural-fortran. 
Michael Nielsen's tutorial on neural networks and deep learning
significantly informed the implementation of neural-fortran.
neural-fortran repository is available at
\url{https://github.com/modern-fortran/neural-fortran}.
The complete source code for this manuscript is available at
\url{https://github.com/milancurcic/neural-fortran-paper}.
\end{acks}

\bibliographystyle{ACM-Reference-Format}
\bibliography{references}

\end{document}